\documentclass[journal]{IEEEtran}
\usepackage[numbers,sort&compress]{natbib}
\usepackage{multirow}
\usepackage{graphicx}
\usepackage{subfigure}

\usepackage{float}
\usepackage{booktabs}
\usepackage{subfigure}
\usepackage[cmex10]{amsmath}
\usepackage{color}
\usepackage{algorithmic}
\usepackage{algorithm}
\usepackage{multirow}
\usepackage[misc]{ifsym}
\usepackage{amsfonts}	
\usepackage{url}
\usepackage{array}
\usepackage{footmisc}
\usepackage{epstopdf}
\newcommand{\xw}[1]{{\color{black} #1}}

\ifCLASSINFOpdf
\else
\fi

\begin{document}
%
\title{A Light Dual-Task Neural Network \\ for Haze Removal}
%
%
%

\author{Yu~Zhang,
        Xinchao~Wang,
        Xiaojun~Bi, 
        Dacheng~Tao,~\IEEEmembership{Fellow,~IEEE}
\thanks{Yu~Zhang is with the College of Information and Communication Engineering, the Harbin Engineering University, Heilongjiang 150001, China,  with the UBTECH Sydney Artificial Intelligence Centre and the School of Information Technologies, the Faculty of Engineering and Information Technologies, the University of Sydney, 6 Cleveland St, Darlington, NSW 2008, and also with the School of Software and Advanced Analytics Institute, University of Technology Sydney, 15 Broadway, Ultimo NSW 2007, Austrilia}
\thanks{Xinchao~Wang is with the Stevens Institute of Technology, Hoboken, New Jersey, 07030, United States.}
\thanks{Xiaojun Bi is with the College of Information and Communication Engineering, the Harbin Engineering University, Heilongjiang 150001, China}
\thanks{D. Tao is with the UBTECH Sydney Artificial Intelligence Centre and the School of Information Technologies, the Faculty of Engineering and Information Technologies, the University of Sydney, 6 Cleveland St, Darlington, NSW 2008, Australia.}
\thanks{This work is supported by the Australian Research Council Project FL-170100117.}
\thanks{ \textcircled{c} 2018 IEEE. Personal use of this material is permitted. Permission from IEEE must be obtained for all other uses, in any current or future media, including
reprinting/republishing this material for advertising or promotional purposes, creating new collective works, for resale or redistribution to servers or lists, or euse of any copyrighted component of this work in other works.}

}

\markboth{Submission to IEEE Signal Processing Letters}%
{Shell \MakeLowercase{\textit{et al.}}: Bare Demo of IEEEtran.cls for IEEE Journals}

\maketitle

\begin{abstract}
Single-image dehazing is a challenging problem due to its ill-posed nature. 
Existing methods rely on a suboptimal two-step approach, where 
an intermediate product like a depth map is estimated, 
based on which the haze-free image is subsequently generated using 
an artificial prior formula.
In this paper, we propose a light dual-task Neural Network called LDTNet 
that restores the haze-free image in one shot. 
We use transmission map estimation as an auxiliary task to assist the main task, haze removal,
in feature extraction and to enhance the generalization of the network. 
In LDTNet, the haze-free image and the transmission map are produced simultaneously. 
As a result, the artificial prior is reduced to the smallest extent. 
Extensive experiments demonstrate that our algorithm achieves superior 
performance against the state-of-the-art methods on both synthetic 
and real-world images.
\end{abstract}


%
\IEEEpeerreviewmaketitle

\section{Introduction}
%
%
%
%
\IEEEPARstart{H}{aze} is a common phenomenon when the light is absorbed and scattered by the turbid medium.
It leads to low visibility and contrast in outdoor scenes. 
Hazy image can be described using the Atmospheric scattering model, 
which was firstly proposed in 1976~\cite{Mccartney1977} 
and widely used by computer vision and graphics community. 
Later, Narasimhan and Nayar~\cite{nayar1999,Narasimhan2003,Narasimhan2002,Narasimhan2001} 
improved the model and re-formulated it as:
\begin{equation}
I(x)=J(x)t(x)+A(1-t(x)),
\end{equation}
where $x$ denotes the pixel locations in the image, 
$I(x)$ demotes the hazy image observed, $J(x)$ demotes the real scene without haze, 
$A$ demotes the atmosphere light, and $t(x)$ demotes the medium transmission. 
The term $t(x)$ can be further written as:\begin{equation}
t(x)=\exp(-\beta d(x)),
\end{equation}
where $d(x)$ is the depth of the scene and $\beta$ indicates the scattering coefficient of the atmosphere.\par 

It is difficult to use hazy images directly for most computer vision tasks like object detection~\cite{wang2014tracking,Shen17DSOD}, tracking~\cite{Wang16a,maksai16CVPR}, pose estimation~\cite{TekinArXiv15,Belagiannis16}, behavior analysis~\cite{Wang14a,maksai17ICCV}, and search~\cite{WangTIP11,Wang11ICME}.
Researchers have thus been devoting great efforts in haze 
removal to restore images with high quality,
among which haze removal from a single image becomes the focus. 
For example, Tan~\cite{Tan2008}, Fattal~\cite{Fattal2008} and He~\cite{He2009} 
implemented single-image dehazing using hand-crafted features, 
upon which the approaches of~\cite{Kratz2009,ancuti2010,He2013,Meng2013}
were proposed. 

All above algorithms, however, rely on 
specific hand-crafted features, which are not able to fully characterize the hazy images.
To this end, recent research has been focused on applying Convolutional Neural Network~(CNN)
that is able to automatically extract features to handle this task. 
DehazeNet~\cite{Cai2016} and MSCNN~\cite{ren2016} design CNNs to estimate the transmission map of the hazy image 
and subsequently use it to estimate atmosphere light. 
Then, the haze-free image is computed using Eq.~(1). 
In AOD-Net~\cite{li2017}, the atmosphere scattering model is re-expressed, 
and the atmosphere light, scattering, and transmission map are rewritten into one matrix. 
AOD-Net estimates this matrix and introduces addition layers as well as multiplication layers 
to compute  the re-expressed formula. 
Although AOD-Net estimates transmission map and atmosphere light at the same time, it first produces an intermediate product, i.e., the matrix of estimated parameters,and then computes the haze-free image by an artificial formula based on the obtained matrix.The errors of the intermediate step may therefore propagate to the dehazing part and thus downgrade the results. 
\xw{GFN~\cite{ren2018gated}, on the other hand, directly estimates the clear scenes from hazy images but relies complex pre-processing operations including white balance, contrast enhancing, and gamma correction. }

In this paper, we propose a light end-to-end dehazing deep network,
termed Light Dual-Task Network~(LDTNet).
In contrast to prior models like AOD-Net that 
decouples the process  into two steps,
ours estimates dehazed images from hazy ones in one shot,
in other words, we do not rely on any intermediate output.
Furthermore, our model does not rely on artificial priors,
such as atmospheric scattering model and the one of~Eq.~(1).
To facilitate the feature learning process, 
we introduce Multitask Learning (MTL) \cite{Caruana1997}
to estimate the haze-free image and transmission map
simultaneously. 

The reason we incorporate the transmission map estimation as an auxiliary task
is that, the features learned for this task can benefit our main task of dehazing.
With a single-task dehazing network, however, it may be very tough to learn such features.
Also, learning the main task alone bears the risk of overfitting~\cite{He2016},
which can be alleviated by the joint learning with the auxiliary one.
In fact, this phenomenon of auxiliary task benefiting the main one
has been observed in a wide domain of 
high-level vision and text tasks~\cite{Sun2014,McLaughlin2017,Ren2015,He2016,Arik2017},
\begin{figure*}[htb]
	\centering 
	\includegraphics[width=0.95\textwidth]{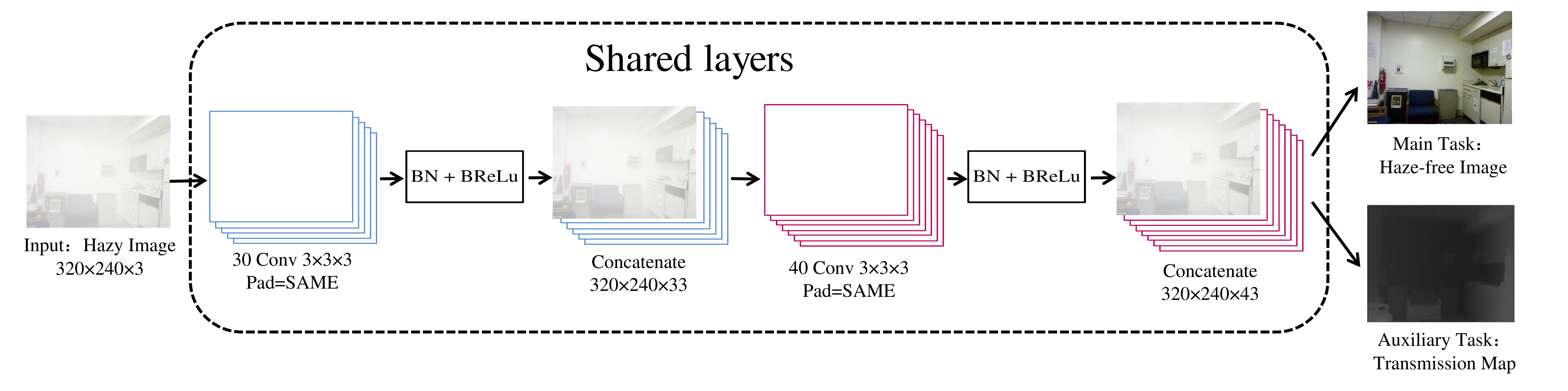}
	\vspace{-0.5cm}
	\caption{The architecture of LDTNet. It takes a hazy image as input, and outputs a dehazed image and an estimated transmission map. } 
	\label{fig1}
\vspace{-0.5cm}
\end{figure*}

Our contribution is therefore a light Multitask dehazing deep network
that does not depend on artificial priors, and that produces 
haze-free image and transmission map simultaneously in one shot.
Our model yields superior results, compared to the state of the art,
on both synthetic and real-world data.


\section{Model}
In this section, we introduce our proposed  LDTNet. 
We start by showing the architecture design and then introduce the loss function.
Unlike prior models that heavily rely on artificial priors, either the hand-crafted features
or hypothetical dehazing models, our approach jointly learns two tasks,
the main task of dehazing and auxiliary one of transmission map estimation,
without human-provided priors in one shot.


 

\subsection{Architecture Design of LDTNet}
The proposed LDTNet, with the help of the auxiliary task, is able to restore haze-free image with a lightweight structure. 
The LDTNet is composed of three cascaded convolutional layers, where the restored image and 
the estimated transmission map are obtained in the third layer. 
We show the architecture of LDTNet in Fig.~1. 

The two tasks share the first two convolutional layers by hard parameter sharing. 
The sizes of convolutional kernels in these two layers are all $3\times3$ and the output channels are 30 and 40 respectively. 
The input three-channels RGB hazy image is concatenated to these two layers severally as three additional feature maps. 
This operation provides  information contained in the input image, ensuring the refinement of the features layer by layer. 
There are two parts of the last layer. 
They combine the feature maps of the 
second convolutional layer in different ways to reconstruct the haze-free image and transmission map respectively. 
The sizes of convolutional kernels in these two parts are all $1\times1$ while the output channels are 3 and 1 respectively.\par 

In LDTNet, no pooling layers are used and zero pixels are padded to the features maps,
ensuring the size of the output image to be consistent with that of the input map.
Furthermore, batch normalization is 
 applied after the first two convolutional layers. Bilateral Rectified Linear Unit~(BRelu)~\cite{Cai2016} is adopted 
 as our activation function.  Specifically,
 BRelu is a modified version of Rectified Linear Unit (Relu)~\cite{Nair2010} with the upper limit set to be 1. 
 It ensures that the pixels in the restored image are constrained in the range of $[0,1]$.

\subsection{Loss Function Design of LDTNet}
In LDTNet,  we simultaneously tackle dehazing and transmission estimation.
We  take the loss function to be
\begin{equation}
L=(1-\alpha)L_{D}(J(x),J^{*}(x))+\alpha L_{T}(t(x),t^{*}(x)),
\end{equation}
where $L_{D}$ and $L_{T}$ correspond to the dehazing loss 
and transmission loss respectively, and $\alpha$ balances the two.
In our implementation, 
both $L_{D}$ and $L_{T}$ take the form of
square loss, defined on the pixel-wise difference
between the ground truth and the prediction of the network.

\vspace{-0.3cm}
\begin{figure}[ht]
	\centering 
	\includegraphics[width=0.28\textwidth]{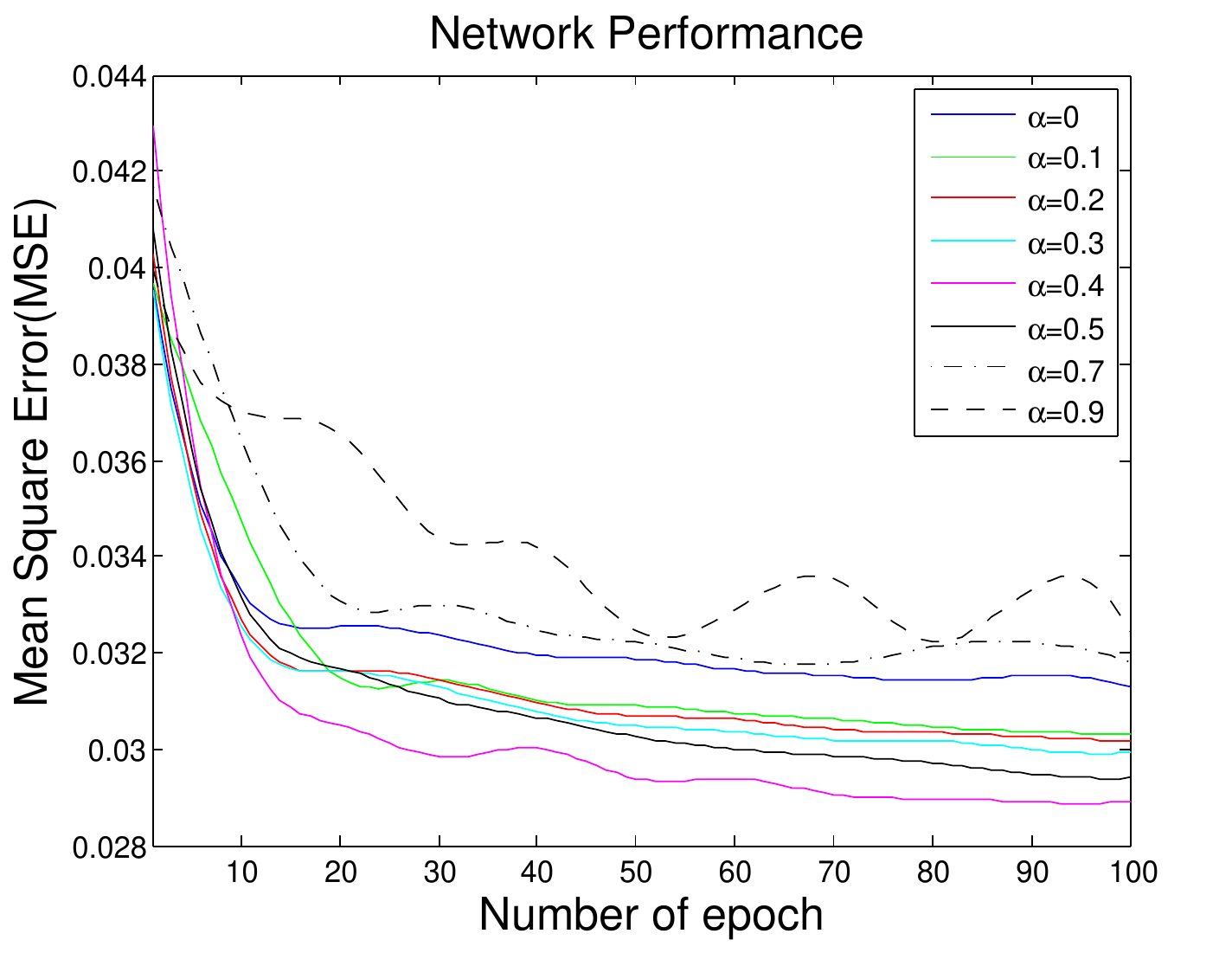}
	\caption{{\color{black}Network performance with different $\alpha$. }
		}  
	\label{fig2}
	\vspace{-0.3cm}
\end{figure}
\vspace{-0.0cm}

\begin{figure*}[htb]
    \centering
    \includegraphics[height=0.42\textwidth,width=0.97\textwidth]{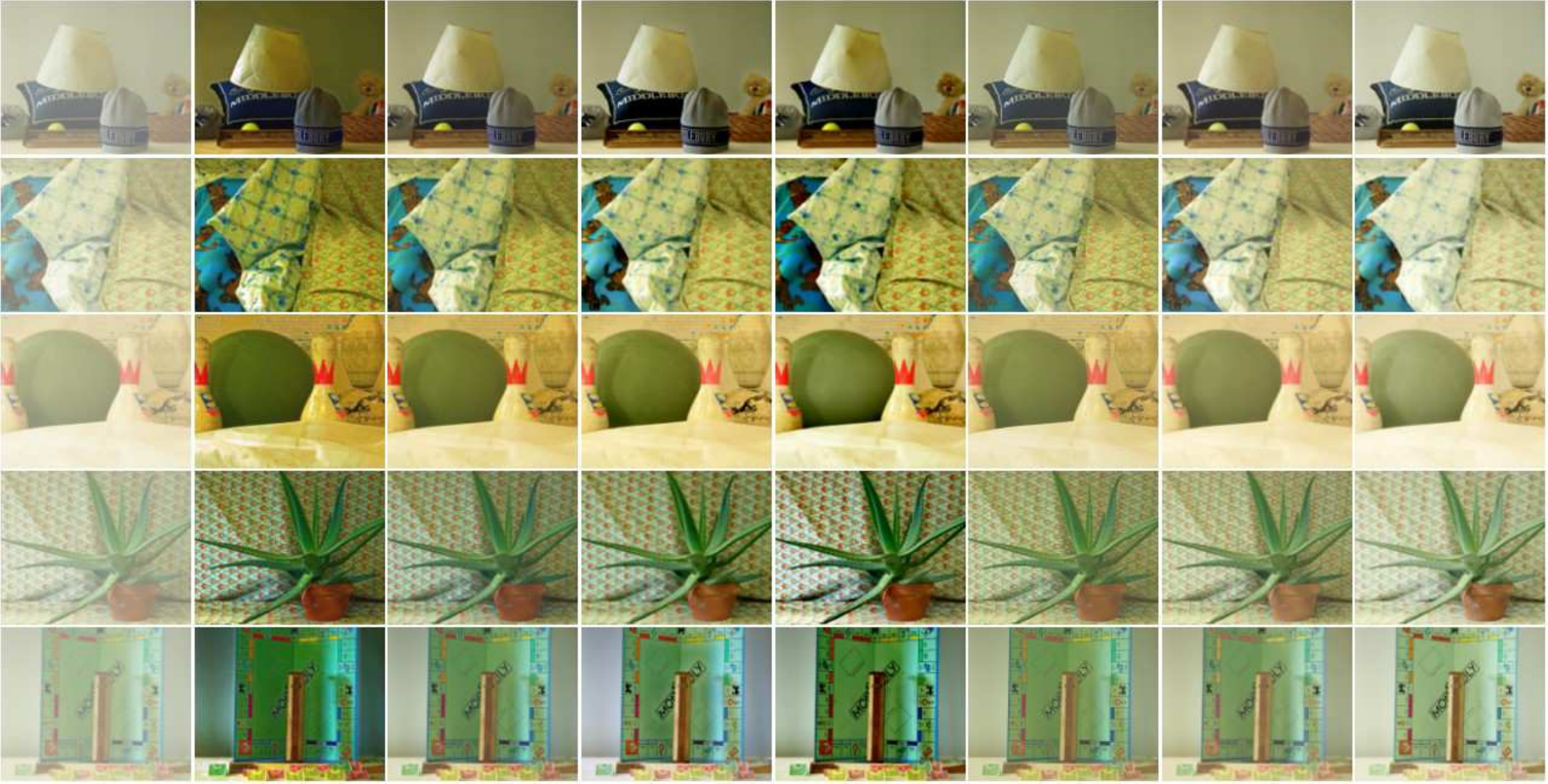}

\footnotesize  \begin{tabular}{p{1.78cm}<{\centering}p{1.78cm}<{\centering}p{1.78cm}<{\centering}p{1.78cm}<{\centering}p{1.78cm}<{\centering}p{1.78cm}<{\centering}p{1.78cm}<{\centering}p{1.78cm}<{\centering}}
	(a)&(b)&(c)&(d)&(e)&(f)&(g)&(h)
\end{tabular}
\caption{Comparative dehazing results  on our synthetic dataset. 
(a)~The haze images,
(b)~DCP,
(c)~CAP,
(d)~DehazeNet,
(e)~MSCNN, 
(f)~OAD-Net,
(g)~LDTNet,
and (h)~Ground truth.}
\label{Fig.3}
\end{figure*}

\section{Experiments}
In this section, we show our experimental validation of the proposed LDTNet.
We first introduce the baselines methods we use for comparison,
and then present our training strategy. Afterwards, we 
show the effectiveness of  our dual-task learning, 
followed by the comparative results on synthetic 
and real-world test images. 
We finally provide the robustness analysis of different methods.

\subsection{Baselines}
We compare our proposed LDTNet with several state-of-the-art methods briefly introduced as follows.
\begin{itemize}
	\item{DCP~\cite{Fattal2008}: {
	The thickness of haze is estimated first and then the haze-free image is recovered using Eq.~(1). }}

	\item{CAP~\cite{Zhu2015}: {Color attenuation prior is utilized for estimating the scene depth,
	 which is further used for computing the haze-free image. }}
	
	\item{MSCNN~\cite{ren2016}: {The transmission map is estimated and refined using two CNNs of different scales,
	which is then used to obtain the haze-free image by Eq.~(1).  } }
	
	\item{DehazeNet~\cite{Cai2016}: {The transmission map is estimated using a CNN
	with a novel nonlinear activation function. } }
	
	\item{AOD-Net~\cite{li2017}: {The transmission map and the airlight are jointly learned using a CNN.} }	
\end{itemize}

\subsection{Training}
To train the LDTNet, we synthesize 10,000 triples of hazy images, haze-free images and transmission maps,
all of which are resized to $240\times320$ pixels based on the  NYU depth~\cite{Silberman2012} dataset using~(1) and~(2). 
We set $ A\in(0.7,1.0) $ and $ \beta\in(0.5,1.5) $, and thus our dataset covers various 
atmosphere situations, multifarious levels of haze, as well as different weather conditions.
For LDTNet, we use Adam~\cite{Kingma2014} to be our optimizer and set the batch size to be~4. 
We implement our model using Tensorflow~\cite{Abadi2015} and the Tensorlayer~\cite{Dong2017} package.
\xw{It takes about 17 hours to train the LDTNet for 100 epochs on a Titan X GPU with Intel~i7 CPU.}

\subsection{Validity of Dual-task Learning}
To validate our dual-task learning, we compute the Mean Square Error~(MSE) 
between our result and ground truth under various values of~$\alpha$ 
on our validation set,  as shown in Fig.~2. 
When $\alpha=0$, 
the auxiliary task is removed and 
only the main task dehazing plays a role,
in which case we obtain the largest MSE. 
This shows that our multitask learning is indeed helpful. 
Also, as can be seen, the performance of network comes to the best when $\alpha=0.4$
on our validation set. 

\begin{figure*}[htb]
	\centering
	\includegraphics[height=0.46\textwidth,width=0.93\textwidth]{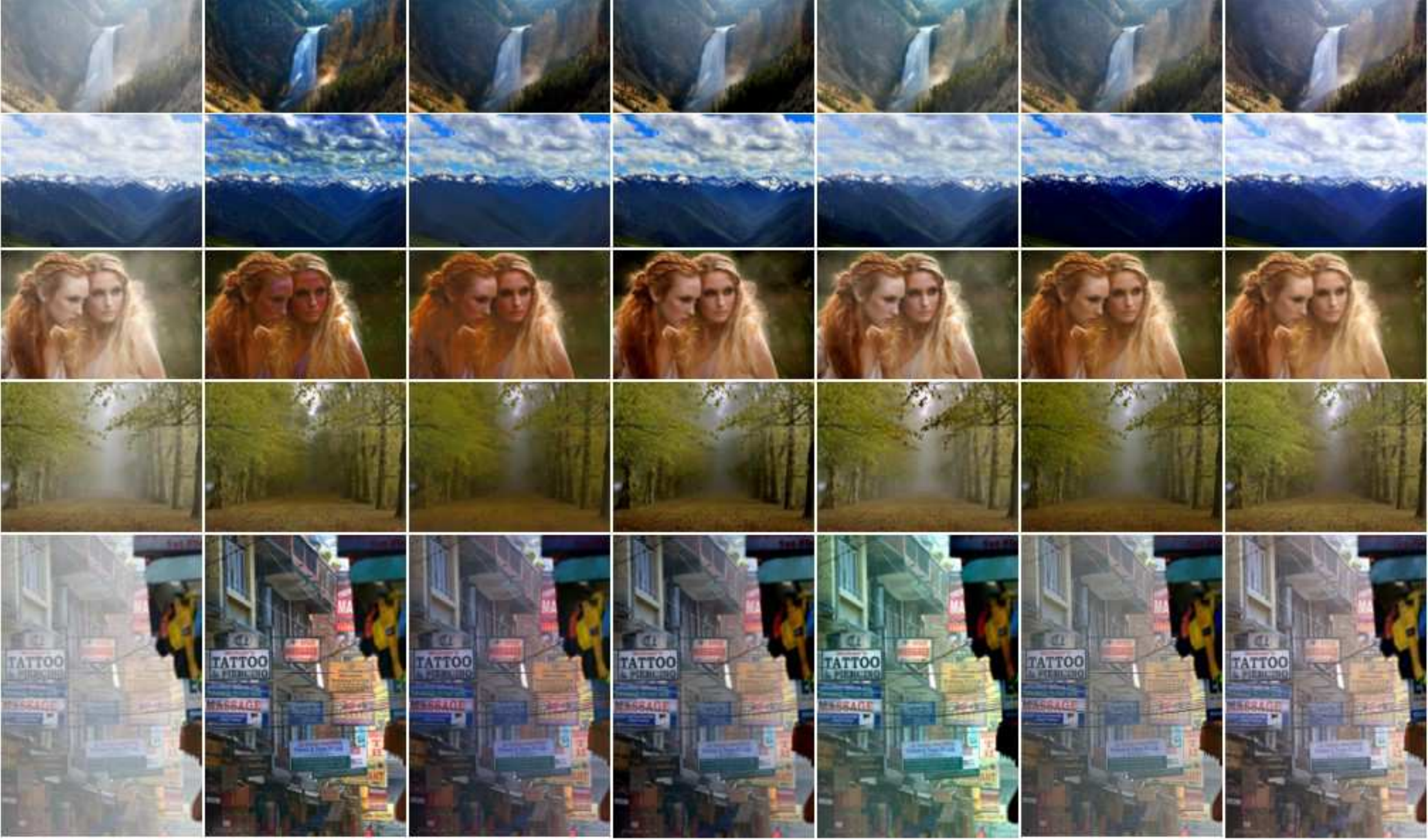}
	\footnotesize  \begin{tabular}{p{2.05cm}<{\centering}p{2.05cm}<{\centering}p{2.05cm}<{\centering}p{2.05cm}<{\centering}p{2.05cm}<{\centering}p{2.05cm}<{\centering}p{2.05cm}<{\centering}}
		(a)&(b)&(c)&(d)&(e)&(f)&(g)
	\end{tabular}
\vspace{-0.2cm}
	\caption{Comparative dehazing results on real-world images. 
		(a)~The hazy images,
		(b)~DCP, (c)~CAP, 
		(d)~DehazeNet, 
		(e)~MSCNN,
		(f)~OAD-Net, 
		and~(g)~LDTNet.}
	\label{Fig.4}
	\vspace{-0.5cm}
\end{figure*}

\subsection{Performance on Synthetic Test Dataset}
We use a synthetic dataset including 21 pairs of stereo images generated using the Middlebury stereo database~\cite{Scharstein2002,Hirschmuller2007,Scharstein2003} to verify the performance of different models. 
The atmosphere light $A$ is set to be~0.85 and the scattering coefficient $\beta$ is set to~1.
These values correspond to the medians of the domains of $A\in(0.7,1)$ and $\beta\in(0.5,1.5)$. 
\xw{It takes about 0.3 second to produce a dehazed image during testing.}

\begin{table}[htbp]
	\centering
	\caption{Average PSNR and SSIM of different dehazing methods on the synthetic dataset}
	\label{table I}
	\begin{tabular}{p{0.7cm}<{\centering} p{0.7cm}<{\centering} p{0.7cm}<{\centering} p{0.8cm}<{\centering} p{1cm}<{\centering} p{1.1cm}<{\centering} p{0.8cm}<{\centering}}
		\hline
		\hline
		Metrics & DCP     & CAP     & MSCNN   & DehazeNet  & AOD-Net & LDTNet  \\ \hline
		PSNR & 14.6713 & 20.9303 & 20.2855 & 21.9690   & 21.3655 & \textbf{24.6156} \\
		SSIM & 0.8432  & 0.9452  & 0.9274  & 0.9463    & 0.9419  & \textbf{0.9517} \\ \hline \hline
	\end{tabular}
\end{table}

We show the obtained mean PSNR and SSIM of the results in Table~I. 
LDTNet achieves the highest PSNR and SSIM scores. 
In Fig.~3, we show the comparative dehazing results of five examples: 
\emph{\textbf{Midd, Cloth, Bowling,Aloe}} and \emph{\textbf{Monopoly}}.


As can be seen,  LDTNet yields visually plausible results which are very similar to the ground truths.
The restored images of the other methods have larger color distortions or over-saturations. 
They are over-sensitive to regions of light colors like white, as they appear similar to haze.
For instance, in the case of~\emph{\textbf{Cloth}}, 
baseline methods tend to produce yellowish colors for the regions that are supposed be of
white color.

Baseline methods fail to produce very accurate results, in part due to
their hand-crafted features or their two-step nature. 
In the cases of DCP and CAP, they rely on a limited set of handcraft features
that may not be expressive enough for some scenes.
MSCNN and DehazeNet, on the other hand, 
first estimate  the intermediate transmission maps and then produce the dehazed results.
The errors occurred in the transmission estimation may thus 
propagate to  dehazing and negatively influence the results. 
Although AOD-Net estimates the transmission map and atmosphere together 
using a CNN, it still relies on an artificial formula to obtain the haze-free image.

By contrast, LDTNet directly learns a mapping from hazy images to the haze-free ones by taking
the transmission map estimation as an auxiliary task. 
As a result, LDTNet does not suffer from the limitations of intermediate products and artificial priors.



\subsection{Performance on Real-world Test Images}
We qualitatively compare our algorithm with DCP, CAP, 
MSCNN, DehazeNet and AOD-Net on~{50} challenging 
real-world images. We show a few of them in Fig. 4. 
As can be seen from the second column, DCP leads to  over enhancement especially on the first three images.
CAP is yields better results than DCP does in terms of over enhancement, but 
fails to preserve textural details in the region of similar colors, like the mountains in the second image
and the hair of the girl in the third.
MSCNN, DehazeNet, and AOD-Net also suffer a certain degree of over enhancement.
In addition, MSCNN causes hue distortion 
in the last image of the fifth column.
The results of LDTNet are indeed more visually plausible, without noticeable color distortions or  loss of details.


\subsection{Robustness Analysis}

We conduct robustness analysis, as done in DehazeNet~\cite{Cai2016}, 
for baselines methods and ours using four types of evaluations, i.e., 
airlight robustness evaluation~(ARE), coefficient robustness evaluation~(CRE), 
scale robustness evaluation~(SRE) and noise robustness evaluation~(NRE).


\begin{table}[]
	
	\centering
	\caption{Average PSNR and SSIM of dehazing results using ARE, CRE, SRE and NRE}
	\label{lable II}
	\begin{tabular}{cccccc}
		
		\hline
		\hline
		&    Metrics & ARE              & CRE              & SRE              & NRE              \\ \hline
		DehazeNet & PSNR & 21.7716          & 21.8800          & 22.0891          & 20.0235          \\
		& SSIM & 0.9450           & 0.9423           & 0.9416           & 0.3469           \\
		MSCNN     & PSNR & 20.1496          & 20.1152          & 20.4302          & 19.8722          \\
		& SSIM & 0.9258           & 0.9240           & 0.9202           & 0.4017           \\
		AOD-Net   & PSNR & 21.1600          & 21.0030          & 21.3820          & 19.9497          \\
		& SSIM & 0.9399           & 0.9395           & 0.9355           & 0.4098           \\
		LDTNet    & PSNR & \textbf{24.1181} & \textbf{24.2780} & \textbf{24.6344} & \textbf{22.2765} \\
		& SSIM & \textbf{0.9459}  & \textbf{0.9468}  & \textbf{0.9441}  & \textbf{0.4925} 
		\\\hline \hline		
	\end{tabular}
\vspace{-0.3cm}
\end{table}
In Table~IV, we show the mean PSNR and SSIM on the  Middlebury stereo dataset.
 In ARE, we synthesize 210 hazy images with $ \beta=1 $ and $ A\in(0.7,1.0) $. 
 Similarly, the same number hazy images are synthesized with $A=0.85$ and $ \beta\in(0.5,1.5) $ for CRE. 
 To analyze the influence of the scale variation,
 we select four scale coefficients, i.e.,  $1,0.8,0.6,0.4$, 
 to generate different scale images with $ \beta=1 $  and $ A=0.85 $. 
Finally, we add three types of noises to the hazy images generated with $ \beta=1 $  and $ A=0.85 $ for NRE. 
The three kinds of noises are Gaussian noise, Poisson noise and salt \& pepper noise.\par 
As can been seen in Table~II, we achieve the best performance in the four types of evaluations. 
This is because our dual-task learning approach improves
the network's capability of task-relevant features extraction and generalization. 
As a result,  LDTNet can extract effective features from the hazy images even in the presence of 
noise and various $A$ and $\beta$. 
Also, since we fuse the original hazy images into the convolutional layers,
the information of the source image can be preserved and utilized to a greater extent. 


\section{Conclusion}
In this paper, we have presented a light dual-task neural network, LDTNet,
which takes as input hazy images and produces dehazed ones in one shot, without
any intermediate results. 
The auxiliary task, transmission map estimation, proves to be helpful for enhancing
dehazing and for improving the network's generalization capability.
We conduct quantitative and qualitative evaluations, and compare our results with 
those of the state-of-the-art methods on both both synthetic and real-world hazy images.
LDTNet yields the most promising results in terms of both accuracy and robustness.
\xw{In our future work, we will explore to  simultaneously conduct dehazing and other tasks,
such as high-level object detection~\cite{Wang2018CVPR} and tracking~\cite{Wang2018TIP1,wang17TIP} ones,
and low-level super-resolution~\cite{Wang2018TIP2} and image restoration~\cite{YinFishEye} ones,
where the both tasks could potentially benefit each other.}

\ifCLASSOPTIONcaptionsoff
\newpage
\fi
{
	\footnotesize	
	\bibliographystyle{ieee}
	\bibliography{ldtnet}

\begin{thebibliography}{46}
\providecommand{\natexlab}[1]{#1}
\providecommand{\url}[1]{#1}
\csname url@samestyle\endcsname
\providecommand{\newblock}{\relax}
\providecommand{\bibinfo}[2]{#2}
\providecommand{\BIBentrySTDinterwordspacing}{\spaceskip=0pt\relax}
\providecommand{\BIBentryALTinterwordstretchfactor}{4}
\providecommand{\BIBentryALTinterwordspacing}{\spaceskip=\fontdimen2\font plus
\BIBentryALTinterwordstretchfactor\fontdimen3\font minus
  \fontdimen4\font\relax}
\providecommand{\BIBforeignlanguage}[2]{{%
\expandafter\ifx\csname l@#1\endcsname\relax
\typeout{** WARNING: IEEEtranN.bst: No hyphenation pattern has been}%
\typeout{** loaded for the language `#1'. Using the pattern for}%
\typeout{** the default language instead.}%
\else
\language=\csname l@#1\endcsname
\fi
#2}}
\providecommand{\BIBdecl}{\relax}
\BIBdecl

\bibitem[Mccartney and Jr(1977)]{Mccartney1977}
E.~J. Mccartney and F.~F.~H. Jr, ``Optics of the atmosphere: Scattering by
  molecules and particles,'' \emph{Optica Acta International Journal of
  Optics}, vol.~14, no.~9, pp. 698--699, 1977.

\bibitem[Nayar and Narasimhan(1999)]{nayar1999}
S.~K. Nayar and S.~G. Narasimhan, ``Vision in bad weather,'' in \emph{Computer
  Vision, 1999. The Proceedings of the Seventh IEEE International Conference
  on}, vol.~2.\hskip 1em plus 0.5em minus 0.4em\relax IEEE, 1999, pp. 820--827.

\bibitem[Narasimhan and Nayar(2003)]{Narasimhan2003}
S.~G. Narasimhan and S.~K. Nayar, ``Contrast restoration of weather degraded
  images,'' \emph{IEEE Transactions on Pattern Analysis \& Machine
  Intelligence}, vol.~25, no.~6, pp. 713--724, 2003.

\bibitem[Narasimhan et~al.(2002)Narasimhan, Srinivasa, Nayar, and
  Shree]{Narasimhan2002}
Narasimhan, G.~Srinivasa, Nayar, and K.~Shree, ``Vision and the atmosphere,''
  \emph{International Journal of Computer Vision}, vol.~48, no.~3, pp.
  233--254, 2002.

\bibitem[Narasimhan and Nayar(2001)]{Narasimhan2001}
S.~G. Narasimhan and S.~K. Nayar, ``Removing weather effects from monochrome
  images,'' in \emph{Computer Vision and Pattern Recognition, 2001. CVPR 2001.
  Proceedings of the 2001 IEEE Computer Society Conference on}, vol.~2.\hskip
  1em plus 0.5em minus 0.4em\relax IEEE, 2001, pp. II--II.

\bibitem[Wang et~al.(2014{\natexlab{a}})Wang, T{\"u}retken, Fleuret, and
  Fua]{wang2014tracking}
X.~Wang, E.~T{\"u}retken, F.~Fleuret, and P.~Fua, ``Tracking interacting
  objects optimally using integer programming,'' \emph{European Conference on
  Computer Vision}, pp. 17--32, 2014.

\bibitem[Shen et~al.(2017)Shen, Shi, Feris, Cao, Yan, Liu, Wang, Xue, and
  Huang]{Shen17DSOD}
Z.~Shen, H.~Shi, R.~Feris, L.~Cao, S.~Yan, D.~Liu, X.~Wang, X.~Xue, and
  T.~Huang, ``{Learning Object Detectors from Scratch with Gated Recurrent
  Feature Pyramids},'' in \emph{{arXiv:1712.00886}}, 2017.

\bibitem[Wang et~al.(2016)Wang, Turetken, Fleuret, and Fua]{Wang16a}
X.~Wang, E.~Turetken, F.~Fleuret, and P.~Fua, ``{Tracking Interacting Objects
  Using Intertwined Flows},'' \emph{IEEE Transactions on Pattern Analysis and
  Machine Intelligence}, vol.~38, no.~11, pp. 2312--2326, 2016.

\bibitem[Maksai et~al.(2016)Maksai, Wang, and Fua]{maksai16CVPR}
A.~Maksai, X.~Wang, and P.~Fua, ``What players do with the ball: A physically
  constrained interaction modeling,'' in \emph{IEEE Conference on Computer
  Vision and Pattern Recognition}, 2016, pp. 972--981.

\bibitem[Tekin et~al.(2015)Tekin, Sun, Wang, Lepetit, and Fua]{TekinArXiv15}
B.~Tekin, X.~Sun, X.~Wang, V.~Lepetit, and P.~Fua, ``{Predicting People's 3D
  Poses from Short Sequences},'' in \emph{{arXiv:1504.08200}}, 2015.

\bibitem[Belagiannis et~al.(2016)Belagiannis, Wang, Shitrit, Hashimoto,
  Stauder, Aoki, Kranzfelder, Schneider, Fua, Ilic, Feussner, and
  Navab]{Belagiannis16}
V.~Belagiannis, X.~Wang, H.~B. Shitrit, K.~Hashimoto, R.~Stauder, Y.~Aoki,
  M.~Kranzfelder, A.~Schneider, P.~Fua, S.~Ilic, H.~Feussner, and N.~Navab,
  ``{Parsing Human Skeletons in an Operating Room},'' \emph{Machine Vision and
  Applications}, vol.~27, no.~7, pp. 1035--1046, 2016.

\bibitem[Wang et~al.(2014{\natexlab{b}})Wang, Ablavsky, BenShitrit, and
  Fua]{Wang14a}
X.~Wang, V.~Ablavsky, H.~BenShitrit, and P.~Fua, ``{Take Your Eyes Off the
  Ball: Improving Ball-Tracking by Focusing on Team Play},'' \emph{Computer
  Vision and Image Understanding}, vol. 119, pp. 102--115, 2014.

\bibitem[Maksai et~al.(2017)Maksai, Wang, Fleuret, and Fua]{maksai17ICCV}
A.~Maksai, X.~Wang, F.~Fleuret, and P.~Fua, ``Non-markovian globally consistent
  multi-object tracking,'' in \emph{IEEE International Conference on Computer
  Vision}, 2017, pp. 2563--2573.

\bibitem[Wang et~al.(2011{\natexlab{a}})Wang, Li, and Tao]{WangTIP11}
X.~Wang, Z.~Li, and D.~Tao, ``{Subspaces indexing model on Grassmann manifold
  for image search},'' \emph{IEEE Transactions on Image Processing}, vol.~20,
  pp. 2627--2635, 2011.

\bibitem[Wang et~al.(2011{\natexlab{b}})Wang, Li, Zhang, and Yuan]{Wang11ICME}
X.~Wang, Z.~Li, L.~Zhang, and J.~Yuan, ``Grassmann hashing for approximate
  nearest neighbor search in high dimensional space,'' in \emph{IEEE
  International Conference on Multimedia and Expo}, 2011, pp. 1--6.

\bibitem[Tan(2008)]{Tan2008}
R.~T. Tan, ``Visibility in bad weather from a single image,'' in \emph{Computer
  Vision and Pattern Recognition, 2008. CVPR 2008. IEEE Conference on}, 2008,
  pp. 1--8.

\bibitem[Fattal(2008)]{Fattal2008}
R.~Fattal, ``Single image dehazing,'' in \emph{Acm Siggraph}, 2008, pp. 1--9.

\bibitem[He et~al.(2009)He, Sun, and Tang]{He2009}
K.~He, J.~Sun, and X.~Tang, ``Single image haze removal using dark channel
  prior,'' in \emph{Computer Vision and Pattern Recognition, 2009. CVPR 2009.
  IEEE Conference on}, 2009, pp. 1956--1963.

\bibitem[Kratz and Nishino(2009)]{Kratz2009}
L.~Kratz and K.~Nishino, ``Factorizing scene albedo and depth from a single
  foggy image,'' in \emph{Computer Vision, 2009 IEEE 12th International
  Conference on}.\hskip 1em plus 0.5em minus 0.4em\relax IEEE, 2009, pp.
  1701--1708.

\bibitem[Ancuti et~al.(2010)Ancuti, Ancuti, Hermans, and Bekaert]{ancuti2010}
C.~O. Ancuti, C.~Ancuti, C.~Hermans, and P.~Bekaert, ``A fast semi-inverse
  approach to detect and remove the haze from a single image,'' in \emph{Asian
  Conference on Computer Vision}.\hskip 1em plus 0.5em minus 0.4em\relax
  Springer, 2010, pp. 501--514.

\bibitem[He et~al.(2013)He, Sun, and Tang]{He2013}
K.~He, J.~Sun, and X.~Tang, ``Guided image filtering,'' \emph{IEEE transactions
  on pattern analysis and machine intelligence}, vol.~35, no.~6, pp.
  1397--1409, 2013.

\bibitem[Meng et~al.(2013)Meng, Wang, Duan, Xiang, and Pan]{Meng2013}
G.~Meng, Y.~Wang, J.~Duan, S.~Xiang, and C.~Pan, ``Efficient image dehazing
  with boundary constraint and contextual regularization,'' in
  \emph{Proceedings of the IEEE international conference on computer vision},
  2013, pp. 617--624.

\bibitem[Cai et~al.(2016)Cai, Xu, Jia, Qing, and Tao]{Cai2016}
B.~Cai, X.~Xu, K.~Jia, C.~Qing, and D.~Tao, ``Dehazenet: An end-to-end system
  for single image haze removal,'' \emph{IEEE Transactions on Image
  Processing}, vol.~25, no.~11, pp. 5187--5198, 2016.

\bibitem[Ren et~al.(2016)Ren, Liu, Zhang, Pan, Cao, and Yang]{ren2016}
W.~Ren, S.~Liu, H.~Zhang, J.~Pan, X.~Cao, and M.-H. Yang, ``Single image
  dehazing via multi-scale convolutional neural networks,'' in \emph{European
  Conference on Computer Vision}.\hskip 1em plus 0.5em minus 0.4em\relax
  Springer, 2016, pp. 154--169.

\bibitem[Li et~al.(2017)Li, Peng, Wang, Xu, and Feng]{li2017}
B.~Li, X.~Peng, Z.~Wang, J.~Xu, and D.~Feng, ``An all-in-one network for
  dehazing and beyond,'' \emph{arXiv:1707.06543}, 2017.

\bibitem[Ren et~al.(2018)Ren, Ma, Zhang, Pan, Cao, Liu, and Yang]{ren2018gated}
W.~Ren, L.~Ma, J.~Zhang, J.~Pan, X.~Cao, W.~Liu, and M.-H. Yang, ``Gated fusion
  network for single image dehazing,'' \emph{arXiv preprint arXiv:1804.00213},
  2018.

\bibitem[Caruana(1997)]{Caruana1997}
R.~Caruana, ``Multitask learning,'' \emph{Machine Learning}, vol.~28, no.~1,
  pp. 41--75, 1997.

\bibitem[He et~al.(2016)He, Huang, Qiao, and Yao]{He2016}
T.~He, W.~Huang, Y.~Qiao, and J.~Yao, ``Text-attentional convolutional neural
  network for scene text detection,'' \emph{IEEE transactions on image
  processing}, vol.~25, no.~6, pp. 2529--2541, 2016.

\bibitem[Sun et~al.(2014)Sun, Wang, and Tang]{Sun2014}
Y.~Sun, X.~Wang, and X.~Tang, ``Deep learning face representation by joint
  identification-verification,'' \emph{Advances in Neural Information
  Processing Systems}, vol.~27, pp. 1988--1996, 2014.

\bibitem[McLaughlin et~al.(2017)McLaughlin, del Rincon, and
  Miller]{McLaughlin2017}
N.~McLaughlin, J.~M. del Rincon, and P.~C. Miller, ``Person reidentification
  using deep convnets with multitask learning,'' \emph{IEEE Transactions on
  Circuits and Systems for Video Technology}, vol.~27, no.~3, pp. 525--539,
  2017.

\bibitem[Ren et~al.(2015)Ren, He, Girshick, and Sun]{Ren2015}
S.~Ren, K.~He, R.~Girshick, and J.~Sun, ``Faster r-cnn: Towards real-time
  object detection with region proposal networks,'' in \emph{Advances in neural
  information processing systems}, 2015, pp. 91--99.

\bibitem[Arik et~al.(2017)Arik, Chrzanowski, Coates, Diamos, Gibiansky, Kang,
  Li, Miller, Raiman, Sengupta, et~al.]{Arik2017}
S.~O. Arik, M.~Chrzanowski, A.~Coates, G.~Diamos, A.~Gibiansky, Y.~Kang, X.~Li,
  J.~Miller, J.~Raiman, S.~Sengupta \emph{et~al.}, ``Deep voice: Real-time
  neural text-to-speech,'' \emph{arXiv preprint arXiv:1702.07825}, 2017.

\bibitem[Nair and Hinton(2010)]{Nair2010}
V.~Nair and G.~E. Hinton, ``Rectified linear units improve restricted boltzmann
  machines,'' in \emph{International Conference on International Conference on
  Machine Learning}, 2010, pp. 807--814.

\bibitem[Zhu et~al.(2015)Zhu, Mai, and Shao]{Zhu2015}
Q.~Zhu, J.~Mai, and L.~Shao, ``A fast single image haze removal algorithm using
  color attenuation prior,'' \emph{IEEE Transactions on Image Processing A
  Publication of the IEEE Signal Processing Society}, vol.~24, no.~11, pp.
  3522--3533, 2015.

\bibitem[Silberman et~al.(2012)Silberman, Hoiem, Kohli, and
  Fergus]{Silberman2012}
N.~Silberman, D.~Hoiem, P.~Kohli, and R.~Fergus, ``Indoor segmentation and
  support inference from rgbd images,'' in \emph{European Conference on
  Computer Vision}, 2012, pp. 746--760.

\bibitem[Kingma and Ba(2014)]{Kingma2014}
D.~Kingma and J.~Ba, ``Adam: A method for stochastic optimization,''
  \emph{Computer Science}, 2014.

\bibitem[Abadi et~al.(2015)Abadi, Agarwal, Barham, Brevdo, Chen, Citro,
  Corrado, Davis, Dean, and Devin]{Abadi2015}
M.~Abadi, A.~Agarwal, P.~Barham, E.~Brevdo, Z.~Chen, C.~Citro, G.~S. Corrado,
  A.~Davis, J.~Dean, and M.~Devin, ``Tensorflow: Large-scale machine learning
  on heterogeneous distributed systems,'' 2015.

\bibitem[Dong et~al.(2017)Dong, Supratak, Mai, Liu, Oehmichen, Yu, and
  Guo]{Dong2017}
H.~Dong, A.~Supratak, L.~Mai, F.~Liu, A.~Oehmichen, S.~Yu, and Y.~Guo,
  ``Tensorlayer: A versatile library for efficient deep learning development,''
  pp. 1201--1204, 2017.

\bibitem[Scharstein et~al.(2002)Scharstein, Szeliski, and
  Zabih]{Scharstein2002}
D.~Scharstein, R.~Szeliski, and R.~Zabih, ``A taxonomy and evaluation of dense
  two-frame stereo correspondence algorithms,'' \emph{International Journal of
  Computer Vision}, vol.~47, no. 1-3, pp. 7--42, 2002.

\bibitem[Hirschmuller and Scharstein(2007)]{Hirschmuller2007}
H.~Hirschmuller and D.~Scharstein, ``Evaluation of cost functions for stereo
  matching,'' in \emph{Computer Vision and Pattern Recognition, 2007. CVPR '07.
  IEEE Conference on}, 2007, pp. 1--8.

\bibitem[Scharstein and Szeliski(2003)]{Scharstein2003}
D.~Scharstein and R.~Szeliski, ``High-accuracy stereo depth maps using
  structured light,'' in \emph{IEEE Computer Society Conference on Computer
  Vision and Pattern Recognition}, 2003, pp. 195--202.

\bibitem[Wang et~al.(2018)Wang, Zhao, Li, Wang, and Tao]{Wang2018CVPR}
F.~Wang, L.~Zhao, X.~Li, X.~Wang, and D.~Tao, ``Geometry-aware scene text
  detection with instance transformation network,'' \emph{IEEE Conference on
  Computer Vision and Pattern Recognition}, 2018.

\bibitem[Lan et~al.(2018)Lan, Wang, Zhang, Tao, Gao, and Huang]{Wang2018TIP1}
L.~Lan, X.~Wang, S.~Zhang, D.~Tao, W.~Gao, and T.~Huang, ``Interacting
  tracklets for multi-object tracking,'' \emph{IEEE Transactions on Image
  Processing}, 2018.

\bibitem[Wang et~al.(2017)Wang, Fan, Chang, Wang, Liu, Tao, and
  Huang]{wang17TIP}
X.~Wang, B.~Fan, S.~Chang, Z.~Wang, X.~Liu, D.~Tao, and T.~Huang, ``Greedy
  batch-based minimum-cost flows for tracking multiple objects,'' \emph{IEEE
  Transactions on Image Processing}, vol.~26, no.~10, pp. 4765--4776, 2017.

\bibitem[Liu et~al.(2018)Liu, Wang, Fan, Liu, Wang, Chang, Wang, and
  Huang]{Wang2018TIP2}
D.~Liu, Z.~Wang, Y.~Fan, X.~Liu, Z.~Wang, S.~Chang, X.~Wang, and T.~Huang,
  ``Learning temporal dynamics for video super-resolution: A deep learning
  approach,'' \emph{IEEE Transactions on Image Processing}, vol.~27, no.~7, pp.
  3432--3445, 2018.

\bibitem[Yin et~al.(2018)Yin, Wang, Yu, Zhang, Fua, and Tao]{YinFishEye}
X.~Yin, X.~Wang, J.~Yu, M.~Zhang, P.~Fua, and D.~Tao, ``{FishEyeRecNet: A
  Multi-Context Collaborative Deep Network for Fisheye Image Rectification},''
  \emph{arXiv:1804.04784}, 2018.

\end{thebibliography}
}
\end{document}